\documentclass{article}
\usepackage{spconf,amsmath,graphicx}

\usepackage{textcomp}

\usepackage{amssymb}
\usepackage{booktabs}
\usepackage{cite}
\usepackage{color}
\usepackage{enumitem}
\usepackage{epsfig}
\usepackage{epstopdf}
\usepackage{gensymb}
\usepackage{latexsym}
\usepackage{multirow}
\usepackage{nonfloat}
\usepackage[all]{nowidow}
\usepackage{pifont}
\usepackage[caption=false,farskip=0pt]{subfig}

\usepackage{hyperref}
\hypersetup{breaklinks=true,colorlinks}

\title{Reflectance-Guided, Contrast-Accumulated Histogram Equalization}
\name{Xiaomeng Wu, Takahito Kawanishi, and Kunio Kashino}
\address{Communication Science Laboratories, NTT Corporation, Japan}

\begin{document}
\hyphenation{Un-sharp un-sharp Re-flectances re-flectances Hy-per-pa-ra-me-ter hy-per-pa-ra-me-ter Hy-per-pa-ra-me-ters hy-per-pa-ra-me-ters Down-sam-pling down-sam-pling Down-sam-pled down-sam-pled Bicu-bic bicu-bic MAT-LAB Cent-OS De-nois-ing de-nois-ing Trade-off trade-off}
\ninept
\maketitle

\begin{abstract}
Existing image enhancement methods fall short of expectations because with them it is difficult to improve global and local image contrast simultaneously. To address this problem, we propose a histogram equalization-based method that adapts to the data-dependent requirements of brightness enhancement and improves the visibility of details without losing the global contrast. This method incorporates the spatial information provided by image context in density estimation for discriminative histogram equalization. To minimize the adverse effect of non-uniform illumination, we propose defining spatial information on the basis of image reflectance estimated with edge preserving smoothing. Our method works particularly well for determining how the background brightness should be adaptively adjusted and for revealing useful image details hidden in the dark.
\end{abstract}


\section{Introduction}
\label{s:introduction}

The aim of image enhancement is to extract hidden image details and enhance the contrast in images with a low dynamic range. The large number of images captured by the widespread use of digital imaging devices has significantly increased the demand for image enhancement. Current commercial raster graphics editors require image editing expertise or considerable manual effort to produce satisfactory results. Therefore, it is important to realize an automated image enhancement technique that adapts to different input image situations. Existing methods of image enhancement can be classified into one of two groups: model-based methods~\cite{JobsonRW97a,AriciDA09,Deng11,LeeLK13,WangZHL13,EilertsenMU15,FuZHZD16,GuoLL17,WuLHK17,WangL18,KinoshitaK19,RenYLL19} and learning-based methods~\cite{GharbiCBHD17,ChenWKC18,HuHXWL18,ParkLYK18,WangZFSZJ19,ZhangZG19}. We concentrate on model-based methods, as they are more interpretable and do not rely on labeled training data.

Histogram equalization (HE) has received the most attention because of its intuitive implementation quality and high efficiency. It aims at deriving an intensity mapping function such that the entropy of the distribution of output intensities can be maximized. Despite its usefulness, HE is wholly indiscriminate; intensities with large pixel populations are expanded to a larger range even if they are of little visual importance. This issue has been addressed by incorporating spatial information in density estimation~\cite{AriciDA09,LeeLK13,EilertsenMU15,WuLHK17}. For example, in contrast-accumulated HE (CACHE)~\cite{WuLHK17}, Wu et al. incorporated image gradient in density estimation to avoid the over-enhancement of background noise by equalizing the reconstructed histogram. Since the spatial information is usually weak for images with non-uniform illumination, the effectiveness of the above methods is limited with respect to brightness and detail enhancement in dark areas.

In Retinex theory, a captured image is taken to be a combination of reflectance and illumination components; the former constitutes the sharp details in the image, while the latter is spatially smooth. Several studies~\cite{JobsonRW97a,GuoLL17} have been conducted where reflectance is considered to be the desired enhancement output and can be obtained by estimating and removing the illumination component. These methods completely eliminate the global contrast, and so may lead to excessively enhanced brightness. Unsharp masking~\cite{Deng11} decomposes an image into high-frequency and low-frequency components, and realizes contrast enhancement by processing the two components individually. In NPE~\cite{WangZHL13}, the illumination is estimated through a bright-pass filter, enhanced with a histogram specification technique, and recombined with the reflectance to reconstruct the desired output. SRIE~\cite{FuZHZD16} adopts a similar practice but enhanced the illumination with gamma correction. These methods, when properly tuned, are effective for discovering dark image details. However, neither histogram specification nor gamma correction takes spatial information into account, and so the global contrast improvement with these methods does not always adapt well to the content of the image.

The objective of this study is to propose an image enhancement method that can simultaneously and adaptively improve global and local contrast. The method should adapt to the data-dependent requirements of brightness enhancement; it should increase the brightness of the area with meaningful image details and leave that area unchanged if not visually significant. For this purpose, we focus on CACHE~\cite{WuLHK17}, as it has a high potential to preserve and improve global contrast. As described above, CACHE incorporates spatial information in density estimation but shows limited effectiveness with respect to detail enhancement in dark areas. We recognize that the main reason for this limitation is that non-uniform illumination significantly reduces the magnitude of the spatial information. To address this issue, we propose defining spatial information on the basis of image reflectance estimated with edge-preserving smoothing such that the HE can be guided by the reflectance for high adaptability. We refer to the proposed method as reflectance-guided CACHE (RG-CACHE). It works particularly well as regards determining how the background brightness should be adjusted and for revealing image details hidden in the dark. Our experiments demonstrate the superiority of RG-CACHE over recent image enhancement methods from both qualitative and quantitative viewpoints.

\section{Proposed Method}
\label{s:method}

\subsection{Overview}
\label{subs:overview}

\begin{figure}[t]
\centering
\includegraphics[width=\linewidth]{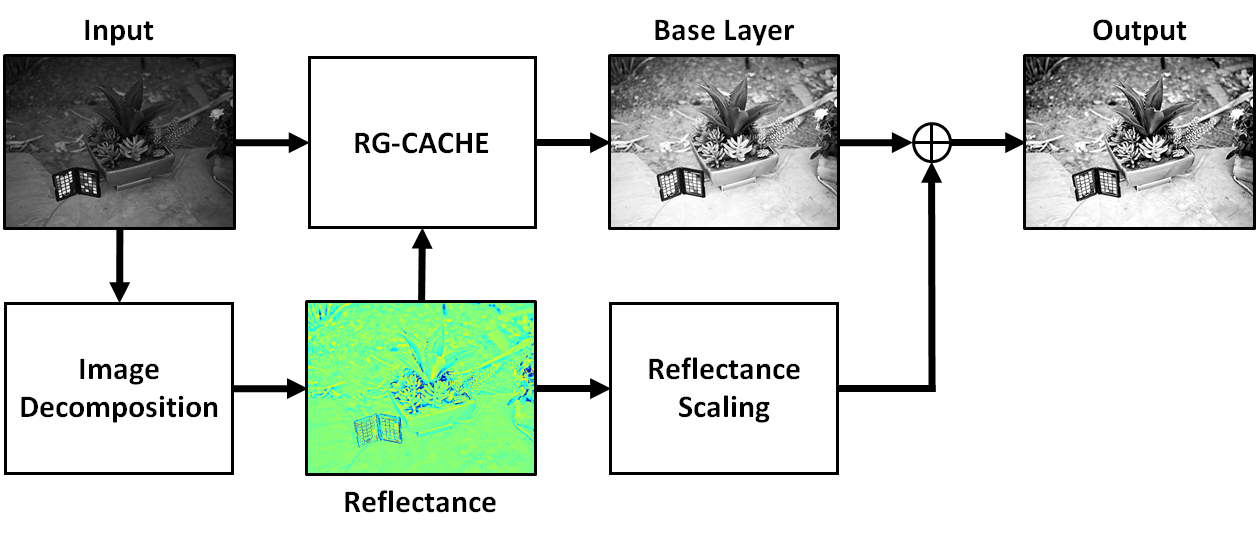}
\caption{Diagram of RG-CACHE for image enhancement.}
\label{f:diagram}
\end{figure}

For a color image $\mathbf{C}_\mathrm{in}$, the lightness $\mathbf{A}_\mathrm{in}$ is defined as the highest of its RGB components, which is equal to the value component of $\mathbf{C}_\mathrm{in}$ in HSV space. The enhanced color image $\mathbf{C}_\mathrm{out}$ is computed by
\begin{equation}
\mathbf{C}_\mathrm{out}=\left(\frac{\mathbf{C}_\mathrm{in}}{\mathbf{A}_\mathrm{in}}\right)\mathbf{A}_\mathrm{out}.
\label{e:cout}
\end{equation}
Here, $\mathbf{A}_\mathrm{out}$ is the lightness after image enhancement. In this paper, all the matrix operations are element-wise operations with implicit expansion enabled. In Fig.~\ref{f:diagram}, the input and the output correspond to $\mathbf{A}_\mathrm{in}$ and $\mathbf{A}_\mathrm{out}$, respectively. Given the input, we obtain its illumination through edge-preserving smoothing and obtain its reflectance by removing the illumination from $\mathbf{A}_\mathrm{in}$ in the logarithmic domain. On one hand, the reflectance is embedded in CACHE, leading to the proposed RG-CACHE. On the other hand, our method allows the user to preserve or enhance local contrast by scaling the reflectance. The base layer provided by RG-CACHE and the scaled reflectance are then combined to output the enhanced lightness $\mathbf{A}_\mathrm{out}$.

\subsection{Contrast-Accumulated Histogram Equalization}
\label{subs:cache}

Let $\mathbf{A}_\mathrm{in}=\{a(q)\}$ where $a(q)\in[0,K)$ is the intensity of each pixel $q$ and $K$ is the total number of intensities (typically 256). Let $n_k$ be the number of occurrences of an intensity $k\in[0,K)$. The probability of an occurrence of a pixel of intensity $k$ in $\mathbf{A}_\mathrm{in}$ is $p_a(k)=n_k/n$, where $n$ is the total number of pixels. We also define the cumulative distribution function (CDF) corresponding to $p_a$ as $P_a(k)$.

We want to find a mapping function of the form $b=T(a)$ to produce an enhanced image $\mathbf{B}=\{b(q)\}$. In HE, the desired $\mathbf{B}$ is assumed to have a flat histogram and a linearized CDF. HE thus chooses the mapping function by using $T(k)=(K-1)P_a(k)$. From this equation, the increment in output intensity vs. a unit step up in input intensity $k-1$, $k\in(0,K)$, can be easily seen to be
\begin{align}
\Delta{T}(k)&=T(k)-T(k-1)\nonumber\\
&=(K-1)p_a(k).
\label{e:deltatk}
\end{align}
Equation~\ref{e:deltatk} indicates that the increment in an intensity, i.e., the contrast gain, is proportional to the probability of the corresponding intensity in the input. This property is usually inconsistent with human vision and is indiscriminate; it may increase the contrast of background noise that has a large pixel population, while reducing the number of usable signals with fewer pixels.

To address this problem, our goal is to help enhance the intensity contrast that is likely to contain usable signals while attenuating it in the case of noise. We incorporate the spatial information of the image into the density estimation such that the goal can be easily achieved by equalizing the reconstructed histogram. More specifically, we reconstruct the probability of an intensity $k$ in $\mathbf{A}_\mathrm{in}$ with
\begin{equation}
\hat{p}_a(k)=\frac{\sum_{q}\varphi(q)\delta(a(q),k)}{\sum_{q}\varphi(q)}.
\label{e:pak}
\end{equation}
Here, $\varphi$ is a spatially varying function expressing the spatial information of each pixel, and $\delta$ is the Kronecker delta. Equation~\ref{e:pak} indicates that each pixel contributes to the density estimation adaptively, and $\hat{p}_a(k)$ can be understood as the expected importance of an intensity given $\mathbf{A}_\mathrm{in}$. Equalizing this equation naturally ensures that the increment $\Delta\hat{T}(k)$ of $k$ is proportional to its expected importance. Our method thus replaces $p_a(k)$ with Eq.~\ref{e:pak} from HE. At stake is how to properly define the spatial information $\varphi$.

\subsection{RG-CACHE}
\label{subs:rg}

\begin{figure}[t]
\centering
\subfloat[$\mathbf{C}_\mathrm{in}$]{
\includegraphics[width=.48\linewidth]{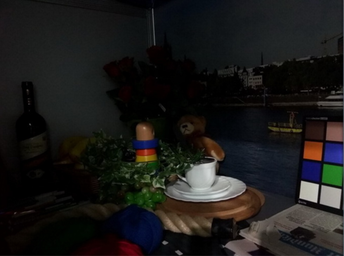}
\label{subf:LIME-8-input}}
\hfill
\subfloat[$\mathbf{C}_\mathrm{out}$ for RG-CACHE]{
\includegraphics[width=.48\linewidth]{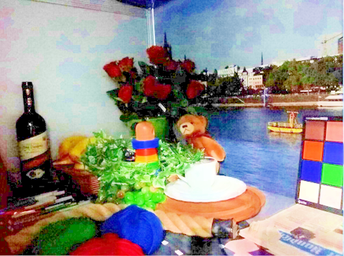}
\label{subf:LIME-8-global-color}}
\\
\subfloat[Illumination $\mathbf{I}$]{
\includegraphics[width=.31\linewidth]{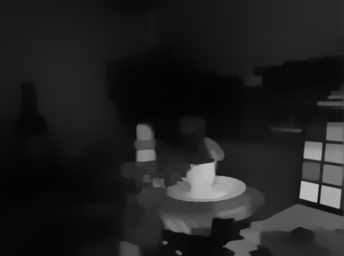}
\label{subf:LIME-8-tsmooth}}
\hfill
\subfloat[Reflectance $\mathbf{R}$]{
\includegraphics[width=.31\linewidth]{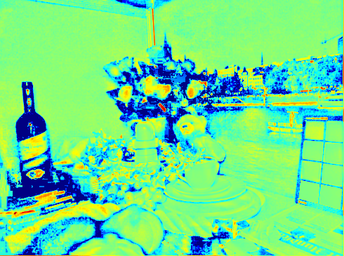}
\label{subf:LIME-8-detail}}
\hfill
\subfloat[$\varphi(q)$]{
\includegraphics[width=.31\linewidth]{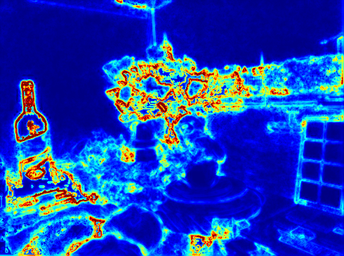}
\label{subf:LIME-8-gradient}}
\\
\subfloat[$\mathbf{C}_\mathrm{out}$ for HE]{
\includegraphics[width=.31\linewidth]{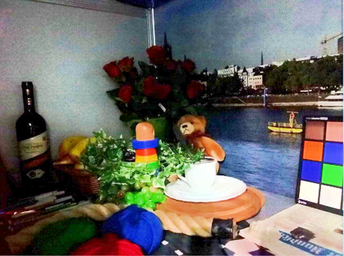}
\label{subf:LIME-8-he}}
\hfill
\subfloat[$\mathbf{C}_\mathrm{out}$ for CACHE]{
\includegraphics[width=.31\linewidth]{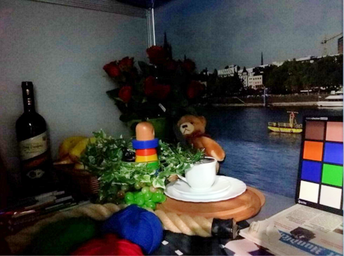}
\label{subf:LIME-8-cache}}
\hfill
\subfloat[$\varphi(q)$ for CACHE]{
\includegraphics[width=.31\linewidth]{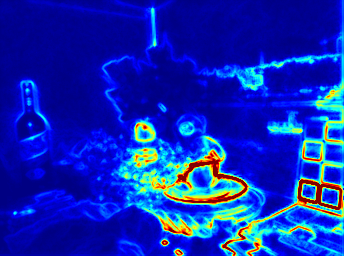}
\label{subf:LIME-8-gradient-cache}}
\\
\subfloat[Histograms with log scale on $x$-axis]{
\includegraphics[height=.31\linewidth]{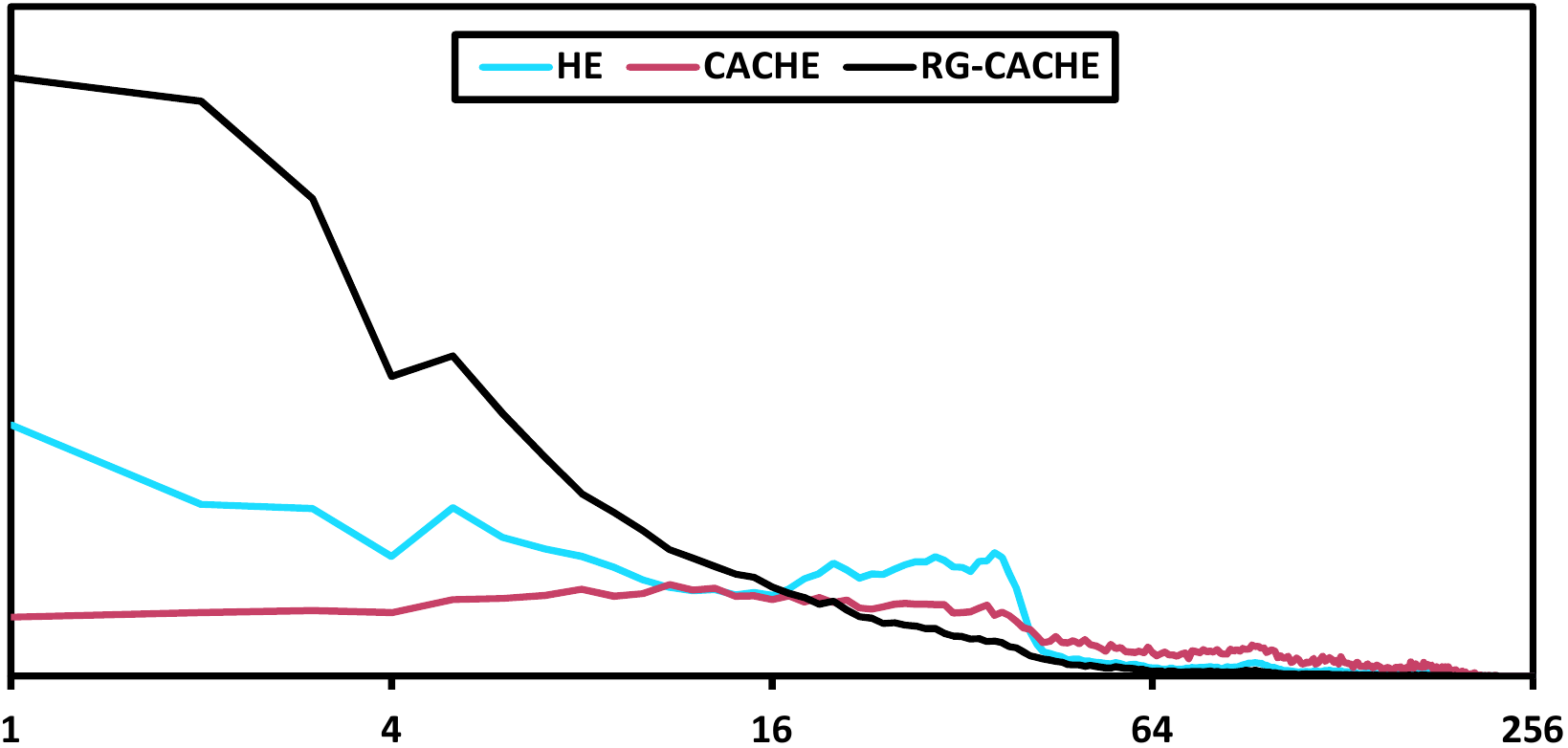}
\label{subf:LIME-8-histogram}}
\hfill
\subfloat[Mapping functions]{
\includegraphics[height=.31\linewidth]{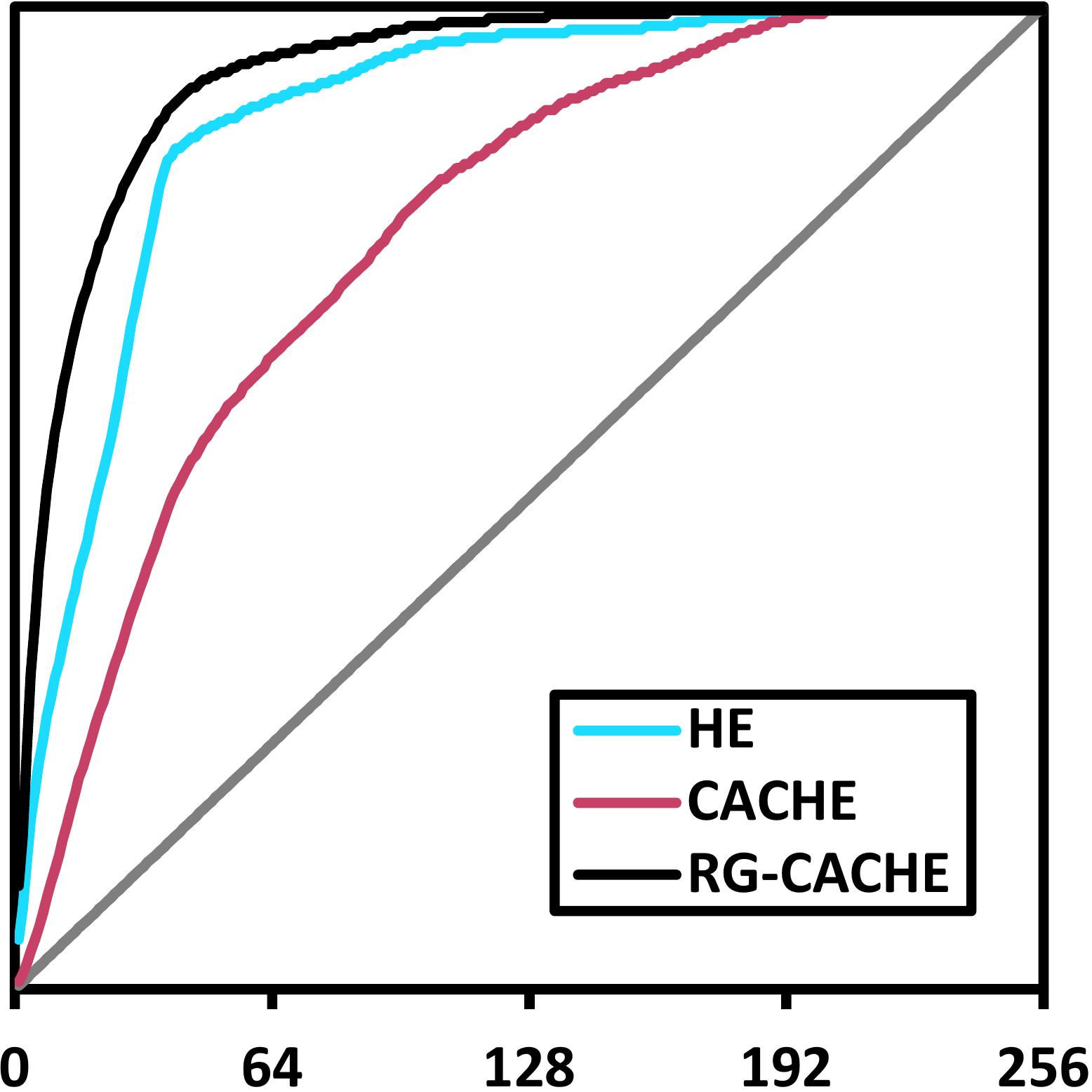}
\label{subf:LIME-8-mapping}}
\caption{Comparison of RG-CACHE with HE and CACHE~\cite{WuLHK17}. \protect\subref{subf:LIME-8-detail}: Positive and negative reflectances are shown in red and blue, respectively, and pixels w/o reflectance are shown in green.}
\label{f:LIME-8}
\end{figure}

$\Delta\hat{T}(k)=(K-1)\hat{p}_a(k)$ is proportional to $\hat{p}_a(k)$, and if we want to increase the brightness of a dark area, we need a large $\hat{p}_a(k)$ for low intensity pixels. One simple solution is to formulate $\varphi(q)$ with each pixel's gradient. Figure~\ref{subf:LIME-8-input} shows an input image containing meaningful objects, e.g., a bunch of flowers, a bottle, and a picture drawn on the wall. Its gradient is shown in Fig.~\ref{subf:LIME-8-gradient-cache}. Despite the wealth of detail in the input, Fig.~\ref{subf:LIME-8-gradient-cache} shows very low gradient values in the dark thus drawing attention to other foreground objects in the light. The main reason for this is that non-uniform illumination significantly reduces the relative magnitude of the gradient in dark areas. Tanaka et al.~\cite{TanakaSO19} suggested that the gradient can be enhanced with an amplification factor that is inversely proportional to the intensity. However, this factor is exclusively dependent on the intensity value and does not take account of the context of the image.

In view of the low gradient being attributable to non-uniform illumination, we propose to eliminate the illumination before defining the function $\varphi(q)$. In this study, we use relative total variation~\cite{XuYXJ12} to decompose $\mathbf{A}_\mathrm{in}$ into illumination and reflectance, but any edge-preserving filter~\cite{XuLXJ11,He0T13,ZhangSXJ14,ZhangXJ14,ShibataTO16,ShenZXJ17} can be used instead. This decomposition is a common technique for tone mapping~\cite{WangZHL13,EilertsenMU15,FuZHZD16}, with which the illumination component is adjusted to enhance the global contrast. Unlike these methods, we focus on the reflectance and allow it to guide the improvement of global contrast based on Eq.~\ref{e:pak}.

Let $\mathbf{I}$ and $\mathbf{R}$ be the illumination and the reflectance components of $\mathbf{A}_\mathrm{in}$, respectively. We define the reflectance by
\begin{equation}
\mathbf{R}=\log_{10}\left(\frac{\mathbf{A}_\mathrm{in}}{\mathbf{I}}\right).
\label{e:r}
\end{equation}
Note that Eq.~\ref{e:r} calculates the reflectance in the logarithmic domain, which is different from related studies~\cite{FuZHZD16,GuoLL17}. It makes our method more effective because logarithmic scaling magnifies the difference in reflectance in the dark area while attenuating the difference in the light area. Let $\mathcal{N}(q)$ denote a set of neighboring coordinates of the pixel $q$. Given $\mathbf{R}=\{r(q)\}$, we compute the gradient with Eq.~\ref{e:varphiq}, where we adopted the von Neumann neighborhood as $\mathcal{N}(q)$.
\begin{equation}
\varphi(q)=\sum_{q'\in\mathcal{N}(q)}|r(q)-r(q')|.
\label{e:varphiq}
\end{equation}

Because natural images contain details on more than one scale, we use a multi-resolution scheme to reliably detect all significant details. We begin by constructing an image pyramid $\mathbf{R}_1,\mathbf{R}_2,\cdots,\mathbf{R}_L$, where $L$ ($=4$ in this paper) is the number of levels. $\mathbf{R}_1=\mathbf{R}$ has the largest resolution and $\mathbf{R}_L$ has the smallest resolution. $\mathbf{R}_2$ is obtained by downsampling $\mathbf{R}_1$ by a factor of two with bicubic interpolation, which is repeated $L-1$ times until $\mathbf{R}_L$ is obtained. At each level, we calculate the gradient according to Eq.~\ref{e:varphiq} from each downsampled $\mathbf{R}$. The function $\varphi$ in Eq.~\ref{e:pak} can thus be calculated by merging the gradients $\varphi$ at all levels using their geometric mean.

Example of the illumination, reflectance, and gradient calculated with our method are shown in Figs.~\ref{subf:LIME-8-tsmooth}, \ref{subf:LIME-8-detail}, and \ref{subf:LIME-8-gradient}, respectively. A comparison of Figs.~\ref{subf:LIME-8-gradient} and \ref{subf:LIME-8-gradient-cache} shows that the proposed reflectance gradient provides a sharper insight into meaningful spatial information hidden in the dark than the conventional gradient. This is also reflected in the histograms $\hat{p}_a(k)$ in Fig.~\ref{subf:LIME-8-histogram} reformulated by the two different types of gradient. RG-CACHE thus produced a mapping function $\hat{T}(k)$ in Fig.~\ref{subf:LIME-8-mapping} that maximizes the brightness enhancement for small intensity values. In areas where there are many objects, but they are buried in the dark, the brightness of background is much better with RG-CACHE (Fig.~\ref{subf:LIME-8-global-color}) than with HE and CACHE~\cite{WuLHK17}.

\subsection{Reflectance Scaling}
\label{subs:rs}

\begin{figure}[t]
\centering
\subfloat[$\mathbf{C}_\mathrm{in}$]{
\includegraphics[width=.31\linewidth]{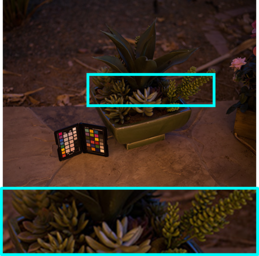}
\label{subf:3-input}}
\hfill
\subfloat[$\mathbf{C}_\mathrm{out}$ ($e=0$)]{
\includegraphics[width=.31\linewidth]{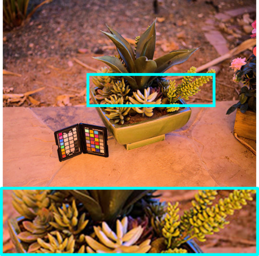}
\label{subf:3-rs00}}
\hfill
\subfloat[$\mathbf{C}_\mathrm{out}$ ($e=0.5$)]{
\includegraphics[width=.31\linewidth]{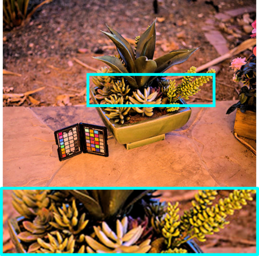}
\label{subf:3-rs05}}
\caption{Reflectance scaling.}
\label{f:reflectance}
\end{figure}

When the input $\mathbf{A}_\mathrm{in}$ is histogram-equalized with RG-CACHE, the result $\mathbf{B}$ can be understood as a base layer and combined with the reflectance component. The image details are enhanced by this combination, since $\mathbf{B}$ already contains the local contrast inherited from $\mathbf{A}_\mathrm{in}$, even if the reflectance is slightly attenuated before the combination. In this study, the output is obtained with $\mathbf{A}_\mathrm{out}=\mathbf{B}+e\mathbf{R}$, where $e$ is a hyperparameter. The effect of $e$ on detail enhancement is shown in Fig.~\ref{f:reflectance}. In our implementation, we empirically set $e=0.5$. Noise-aware detail control~\cite{EilertsenMU15,SuJ17} can also be employed here, but it is not taken into account as it is outside the scope of this paper.

\section{Experiments}
\label{s:experiment}

Six datasets comprising a total of 677 images, including USC-SIPI~\cite{usc}, BSDS500~\cite{ArbelaezMFM11}, LIME~\cite{GuoLL17}, NPE~\cite{WangZHL13}, Vonikakis~\cite{vonikakis}, and Image Compression~\cite{imagecompression}, were used for evaluation. We compared RG-CACHE with HE, CACHE~\cite{WuLHK17}, LIME~\cite{GuoLL17}, and NPIE~\cite{WangL18}. All the implementations were provided by the original authors and are in MATLAB. The experiments were conducted on a machine running CentOS 6.3 with a 160 GB memory and a 3.0 GHz CPU.

\subsection{Qualitative Evaluation}
\label{subs:qualitative}

\begin{figure*}[t]
\centering
\subfloat[Input]{
\includegraphics[width=.15\linewidth]{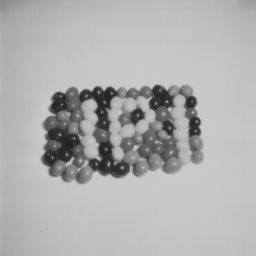}
\label{subf:4107-input}}
\hfill
\subfloat[HE]{
\includegraphics[width=.15\linewidth]{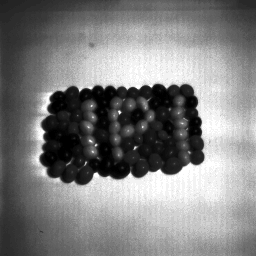}
\label{subf:4107-he}}
\hfill
\subfloat[CACHE~\cite{WuLHK17}]{
\includegraphics[width=.15\linewidth]{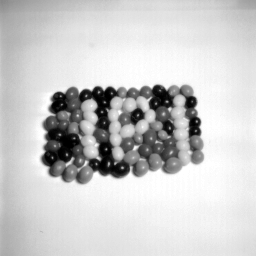}
\label{subf:4107-cache}}
\hfill
\subfloat[LIME~\cite{GuoLL17}]{
\includegraphics[width=.15\linewidth]{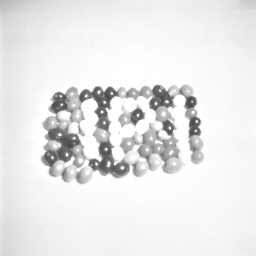}
\label{subf:4107-lime}}
\hfill
\subfloat[NPIE~\cite{WangL18}]{
\includegraphics[width=.15\linewidth]{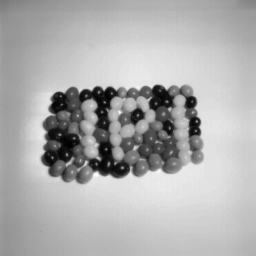}
\label{subf:4107-npie}}
\hfill
\subfloat[RG-CACHE w/ RS]{
\includegraphics[width=.15\linewidth]{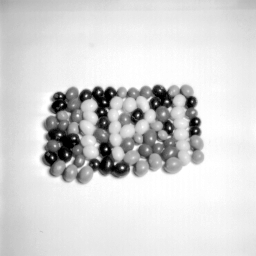}
\label{subf:4107-output}}
\\
\subfloat[Input]{
\includegraphics[width=.15\linewidth]{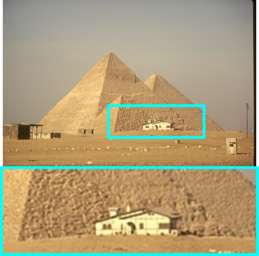}
\label{subf:161062-input}}
\hfill
\subfloat[HE]{
\includegraphics[width=.15\linewidth]{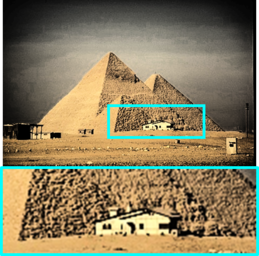}
\label{subf:161062-he}}
\hfill
\subfloat[CACHE~\cite{WuLHK17}]{
\includegraphics[width=.15\linewidth]{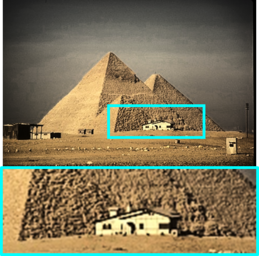}
\label{subf:161062-cache}}
\hfill
\subfloat[LIME~\cite{GuoLL17}]{
\includegraphics[width=.15\linewidth]{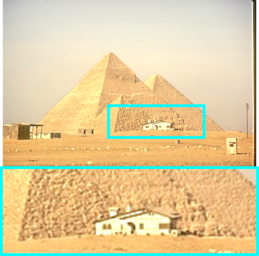}
\label{subf:161062-lime}}
\hfill
\subfloat[NPIE~\cite{WangL18}]{
\includegraphics[width=.15\linewidth]{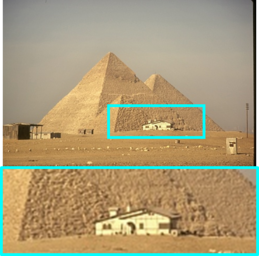}
\label{subf:161062-npie}}
\hfill
\subfloat[RG-CACHE w/ RS]{
\includegraphics[width=.15\linewidth]{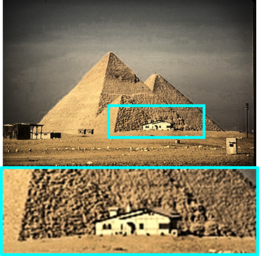}
\label{subf:161062-output}}
\\
\subfloat[Input]{
\includegraphics[width=.15\linewidth]{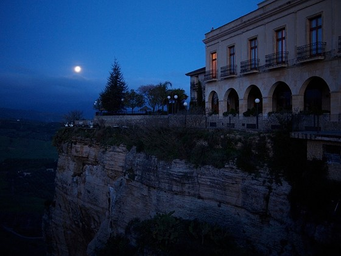}
\label{subf:2-input}}
\hfill
\subfloat[HE]{
\includegraphics[width=.15\linewidth]{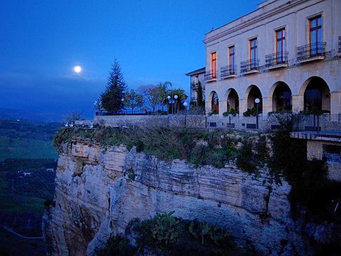}
\label{subf:2-he}}
\hfill
\subfloat[CACHE~\cite{WuLHK17}]{
\includegraphics[width=.15\linewidth]{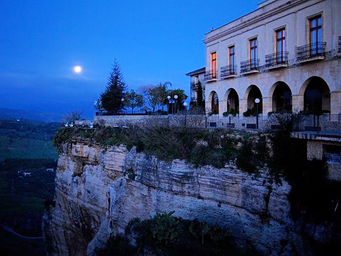}
\label{subf:2-cache}}
\hfill
\subfloat[LIME~\cite{GuoLL17}]{
\includegraphics[width=.15\linewidth]{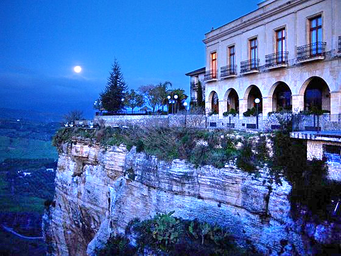}
\label{subf:2-lime}}
\hfill
\subfloat[NPIE~\cite{WangL18}]{
\includegraphics[width=.15\linewidth]{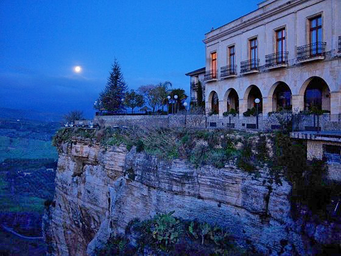}
\label{subf:2-npie}}
\hfill
\subfloat[RG-CACHE w/ RS]{
\includegraphics[width=.15\linewidth]{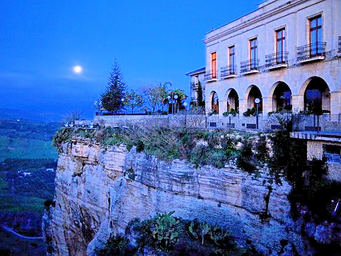}
\label{subf:2-output}}
\\
\subfloat[Input]{
\includegraphics[width=.15\linewidth]{8-input.png}
\label{subf:8-input}}
\hfill
\subfloat[HE]{
\includegraphics[width=.15\linewidth]{8-he.png}
\label{subf:8-he}}
\hfill
\subfloat[CACHE~\cite{WuLHK17}]{
\includegraphics[width=.15\linewidth]{8-cache.png}
\label{subf:8-cache}}
\hfill
\subfloat[LIME~\cite{GuoLL17}]{
\includegraphics[width=.15\linewidth]{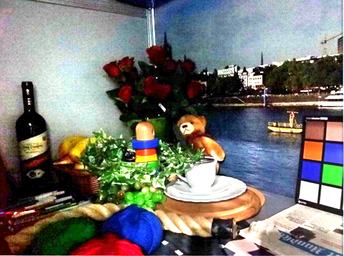}
\label{subf:8-lime}}
\hfill
\subfloat[NPIE~\cite{WangL18}]{
\includegraphics[width=.15\linewidth]{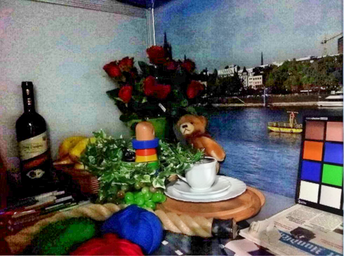}
\label{subf:8-npie}}
\hfill
\subfloat[RG-CACHE w/ RS]{
\includegraphics[width=.15\linewidth]{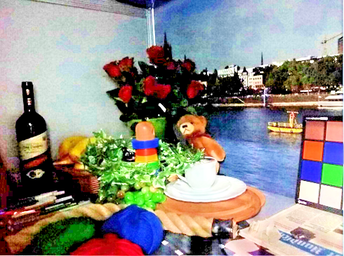}
\label{subf:8-output}}
\caption{Comparison of RG-CACHE w/ RS with state-of-the-art image enhancement (see supplementary material~\cite{appendix} for more examples).}
\label{f:comparison}
\end{figure*}

Examples of the contrast enhancement results are shown in Fig.~\ref{f:comparison}. HE increased the contrast of the meaningless background with high intensity populations while reducing the visibility of foreground objects (Fig.~\ref{subf:4107-he}). CACHE has made natural contrast enhancement possible, but in dark areas the ability to deal with brightness and local contrast is limited. LIME assumes the reflectance component of the image to be the ideal form of contrast enhancement, and eliminates the illumination completely. This can result in the excessive enhancement of the brightness for non-low key lighting images, as in Figs.~\ref{subf:4107-lime} and \ref{subf:161062-lime}, which also results in the loss of local contrast. NPIE is a multilayer extension of NPE~\cite{WangZHL13} where the illumination is estimated through a bright-pass filter, enhanced with histogram specification, and recombined with the reflectance to reconstruct the desired output. Since the histogram specification does not take spatial information into consideration, and because of the nature of the bright-pass filter, NPIE produced somewhat insufficient brightness, as in Figs.~\ref{subf:4107-npie}, \ref{subf:161062-npie}, and \ref{subf:2-npie}. In general, RG-CACHE enabled consistently balanced and natural contrast enhancement. Compared with HE and CACHE, it is particularly effective at improving brightness in dark areas, as in Figs.~\ref{subf:2-output} and \ref{subf:8-output}, where local contrast cannot be easily observed from the input. Meanwhile, it correctly adapted to the non-low key lighting situation of the input, as in Figs.~\ref{subf:4107-output} and \ref{subf:161062-output}, and it avoids the over-enhancement of brightness. (See supplementary material~\cite{appendix} for more examples.)

\subsection{Quantitative Evaluation}
\label{subs:quantitative}

We evaluated the contrast enhancement performance objectively using four metrics, following the practice of previous studies~\cite{LeeLK13,RenYLL19}.

\noindent
\textbf{Discrete entropy (DE).} DE measures the amount of information in an image; a high DE indicates that the image contains more variations and is assumed to have better visual quality.

\noindent
\textbf{Measure of enhancement (EME)~\cite{AgaianSP07}.} EME approximates the average local contrast in an image; a high EME is assumed to indicate greater effectiveness in contrast enhancement.

\noindent
\textbf{PixDist (PD)~\cite{ChenAPA06}.} PD computes the average intensity difference over all the pixel pairs in an image; it yields a high score, indicating better quality, when the pixel intensities are uniformly distributed.

\noindent
\textbf{Patch-based contrast quality index (PCQI)~\cite{WangMYWL15}.} PCQI measures the distortions of contrast strength and structure between input and output; a higher score implies better contrast enhancement.

\begin{table}[t]
\centering
\caption{Quantitative evaluation. The higher the statistics, the better the quality for all the four quality metrics. RS: Reflectance scaling. The best results per column are shown in bold.}
\label{t:quantitative}
\small
\begin{tabular*}{\linewidth}{@{\extracolsep{\fill}}lcccc}
\toprule
\textbf{Method} & \textbf{DE} & \textbf{EME} & \textbf{PD} & \textbf{PCQI} \\
\midrule
No Enhancement & 7.17 & 15.7 & 27.9 & 1.00 \\
\midrule
HE & 7.60 & 36.6 & \textbf{39.3} & 1.05 \\
CACHE~\cite{WuLHK17} & 7.62 & 34.8 & 38.5 & 1.08 \\
LIME~\cite{GuoLL17} & 7.08 & 13.0 & 27.6 & 0.88 \\
NPIE~\cite{WangL18} & 7.33 & 17.4 & 27.8 & 0.99 \\
$\star$ RG-CACHE w/ RS & \textbf{7.65} & \textbf{50.9} & \textbf{39.3} & \textbf{1.09} \\
\bottomrule
\end{tabular*}
\end{table}

Table~\ref{t:quantitative} shows the quality scores of the compared methods, averaged over 500 test images in BSDS500. RG-CACHE achieved the highest scores among all methods. This demonstrates the consistent effectiveness of our method in terms of contrast enhancement. LIME got the worst scores because it completely removes the illumination and reduces contrast strength. NPIE did not show high statistics because the illumination component is compressed with histogram specification to preserve the naturalness of the image.

\subsection{Efficiency}
\label{subf:efficiency}

We evaluated the time required by the compared methods to process the 500 test images of BSDS500. The resolution of all the images was $481\times321$. On average, HE and CACHE took 11 ms and 21 ms, respectively, which were the fastest methods in our experiments. LIME took 176 ms, followed by RG-CACHE 495 ms. The NPIE processing time was 10,494 ms, which is believed to be due to the unoptimized implementation and the nature of the bright-pass filter.

Edge-preserving smoothing was the most time-consuming with RG-CACHE, taking 473 ms. The other processes took only 22 ms, which is equivalent to the time taken by CACHE. We adopted relative total variation~\cite{XuYXJ12} as a filter for edge-preserving smoothing because of its high pixel-level accuracy in the detection and preservation of low-frequency components. If high pixel-level accuracy is non-essential, the processing time of RG-CACHE can be reduced by adopting a much faster filter, as used in previous studies~\cite{EilertsenMU15,GuoLL17}.

\section{Conclusion}
\label{s:conclusion}

In this study, we proposed a novel image enhancement method, RG-CACHE, which resulted in a desirable relationship between spatial information regarding image content and brightness improvement. We showed that the reflectance and the reflectance gradient, obtained by removing the illumination, provide a sharp insight into meaningful spatial information hidden in the dark. This spatial information can be used as an important clue to derive the contrast gain of the intensity, and the brightness of a low-key lighting image can be effectively improved without losing global contrast.

Many high-end cameras provide noise levels below the perceivable contrast of human vision, but after removing the illumination, noise in the reflectance is often pushed into the visible range. The revealed noise enhances the gradient response, which induces RG-CACHE to further amplify the noise. This problem can be mitigated by using an off-the-shelf denoising method~\cite{DabovFKE07,LiuTO13,LiuTO14} or incorporating noise-aware detail control~\cite{EilertsenMU15,SuJ17} into RG-CACHE. We recognize this as our future work. Meanwhile, the HE used by RG-CACHE naturally imposes a monotonicity constraint on the mapping function $\hat{T}(k)$, which may not be desirable for certain applications. We aim to find a solution that allows us to trade off monotonicity for more flexible contrast reproduction.

\bibliographystyle{MyIEEEbib}
\bibliography{refs}

\begin{thebibliography}{10}

\bibitem{JobsonRW97a}
D.~J. Jobson, Z.~Rahman, and G.~A. Woodell,
\newblock ``A multiscale {Retinex} for bridging the gap between color images
  and the human observation of scenes,''
\newblock {\em {IEEE} Trans. Image Processing}, vol. 6, no. 7, pp. 965--976,
  1997.

\bibitem{AriciDA09}
T.~Arici, S.~Dikbas, and Y.~Altunbasak,
\newblock ``A histogram modification framework and its application for image
  contrast enhancement,''
\newblock {\em {IEEE} Trans. Image Processing}, vol. 18, no. 9, pp. 1921--1935,
  2009.

\bibitem{Deng11}
G.~Deng,
\newblock ``A generalized unsharp masking algorithm,''
\newblock {\em {IEEE} Trans. Image Processing}, vol. 20, no. 5, pp. 1249--1261,
  2011.

\bibitem{LeeLK13}
C.~Lee, C.~Lee, and C.~Kim,
\newblock ``Contrast enhancement based on layered difference representation of
  {2D} histograms,''
\newblock {\em {IEEE} Trans. Image Processing}, vol. 22, no. 12, pp.
  5372--5384, 2013.

\bibitem{WangZHL13}
S.~Wang, J.~Zheng, H.~Hu, and B.~Li,
\newblock ``Naturalness preserved enhancement algorithm for non-uniform
  illumination images,''
\newblock {\em {IEEE} Trans. Image Processing}, vol. 22, no. 9, pp. 3538--3548,
  2013.

\bibitem{EilertsenMU15}
G.~Eilertsen, R.~K. Mantiuk, and J.~Unger,
\newblock ``Real-time noise-aware tone mapping,''
\newblock {\em {ACM} Trans. Graph.}, vol. 34, no. 6, pp. 198:1--198:15, 2015.

\bibitem{FuZHZD16}
X.~Fu, D.~Zeng, Y.~Huang, X.~Zhang, and X.~Ding,
\newblock ``A weighted variational model for simultaneous reflectance and
  illumination estimation,''
\newblock in {\em CVPR}, 2016, pp. 2782--2790.

\bibitem{GuoLL17}
X.~Guo, Y.~Li, and H.~Ling,
\newblock ``{LIME:} {Low}-light image enhancement via illumination map
  estimation,''
\newblock {\em {IEEE} Trans. Image Processing}, vol. 26, no. 2, pp. 982--993,
  2017.

\bibitem{WuLHK17}
X.~Wu, X.~Liu, K.~Hiramatsu, and K.~Kashino,
\newblock ``Contrast-accumulated histogram equalization for image
  enhancement,''
\newblock in {\em ICIP}, 2017, pp. 3190--3194.

\bibitem{WangL18}
S.~Wang and G.~Luo,
\newblock ``Naturalness preserved image enhancement using a priori multi-layer
  lightness statistics,''
\newblock {\em {IEEE} Trans. Image Processing}, vol. 27, no. 2, pp. 938--948,
  2018.

\bibitem{KinoshitaK19}
Y.~Kinoshita and H.~Kiya,
\newblock ``Scene segmentation-based luminance adjustment for multi-exposure
  image fusion,''
\newblock {\em {IEEE} Trans. Image Processing}, vol. 28, no. 8, pp. 4101--4116,
  2019.

\bibitem{RenYLL19}
Y.~Ren, Z.~Ying, T.~H. Li, and G.~Li,
\newblock ``{LECARM:} {Low}-light image enhancement using the camera response
  model,''
\newblock {\em {IEEE} Trans. Circuits Syst. Video Techn.}, vol. 29, no. 4, pp.
  968--981, 2019.

\bibitem{GharbiCBHD17}
M.~Gharbi, J.~Chen, J.~T. Barron, S.~W. Hasinoff, and F.~Durand,
\newblock ``Deep bilateral learning for real-time image enhancement,''
\newblock {\em {ACM} Trans. Graph.}, vol. 36, no. 4, pp. 118:1--118:12, 2017.

\bibitem{ChenWKC18}
Y.~Chen, Y.~Wang, M.~Kao, and Y.~Chuang,
\newblock ``Deep photo enhancer: Unpaired learning for image enhancement from
  photographs with {GANs},''
\newblock in {\em CVPR}, 2018, pp. 6306--6314.

\bibitem{HuHXWL18}
Y.~Hu, H.~He, C.~Xu, B.~Wang, and S.~Lin,
\newblock ``Exposure: {A} white-box photo post-processing framework,''
\newblock {\em {ACM} Trans. Graph.}, vol. 37, no. 2, pp. 26:1--26:17, 2018.

\bibitem{ParkLYK18}
J.~Park, J.~Lee, D.~Yoo, and I.~S. Kweon,
\newblock ``Distort-and-recover: Color enhancement using deep reinforcement
  learning,''
\newblock in {\em CVPR}, 2018, pp. 5928--5936.

\bibitem{WangZFSZJ19}
R.~Wang, Q.~Zhang, C.~Fu, et~al.,
\newblock ``Underexposed photo enhancement using deep illumination
  estimation,''
\newblock in {\em CVPR}, 2019, pp. 6849--6857.

\bibitem{ZhangZG19}
Y.~Zhang, J.~Zhang, and X.~Guo,
\newblock ``Kindling the darkness: {A} practical low-light image enhancer,''
\newblock {\em {ACM} Multimedia}, 2019.

\bibitem{TanakaSO19}
M.~Tanaka, T.~Shibata, and M.~Okutomi,
\newblock ``Gradient-based low-light image enhancement,''
\newblock in {\em ICCE}, 2019, pp. 1--2.

\bibitem{XuYXJ12}
L.~Xu, Q.~Yan, Y.~Xia, and J.~Jia,
\newblock ``Structure extraction from texture via relative total variation,''
\newblock {\em {ACM} Trans. Graph.}, vol. 31, no. 6, pp. 139:1--139:10, 2012.

\bibitem{XuLXJ11}
L.~Xu, C.~Lu, Y.~Xu, and J.~Jia,
\newblock ``Image smoothing via \emph{L}\({}_{\mbox{0}}\) gradient
  minimization,''
\newblock {\em {ACM} Trans. Graph.}, vol. 30, no. 6, pp. 174, 2011.

\bibitem{He0T13}
K.~He, J.~Sun, and X.~Tang,
\newblock ``Guided image filtering,''
\newblock {\em {IEEE} Trans. Pattern Anal. Mach. Intell.}, vol. 35, no. 6, pp.
  1397--1409, 2013.

\bibitem{ZhangSXJ14}
Q.~Zhang, X.~Shen, L.~Xu, and J.~Jia,
\newblock ``Rolling guidance filter,''
\newblock in {\em ECCV}, 2014, pp. 815--830.

\bibitem{ZhangXJ14}
Q.~Zhang, L.~Xu, and J.~Jia,
\newblock ``100+ times faster weighted median filter {(WMF)},''
\newblock in {\em CVPR}, 2014, pp. 2830--2837.

\bibitem{ShibataTO16}
T.~Shibata, M.~Tanaka, and M.~Okutomi,
\newblock ``Gradient-domain image reconstruction framework with intensity-range
  and base-structure constraints,''
\newblock in {\em CVPR}, 2016, pp. 2745--2753.

\bibitem{ShenZXJ17}
X.~Shen, C.~Zhou, L.~Xu, and J.~Jia,
\newblock ``Mutual-structure for joint filtering,''
\newblock {\em International Journal of Computer Vision}, vol. 125, no. 1-3,
  pp. 19--33, 2017.

\bibitem{SuJ17}
H.~Su and C.~Jung,
\newblock ``Low light image enhancement based on two-step noise suppression,''
\newblock in {\em ICASSP}, 2017, pp. 1977--1981.

\bibitem{usc}
\url{sipi.usc.edu/database}.

\bibitem{ArbelaezMFM11}
P.~Arbelaez, M.~Maire, C.~C. Fowlkes, and J.~Malik,
\newblock ``Contour detection and hierarchical image segmentation,''
\newblock {\em {IEEE} Trans. Pattern Anal. Mach. Intell.}, vol. 33, no. 5, pp.
  898--916, 2011.

\bibitem{vonikakis}
\url{sites.google.com/site/vonikakis/datasets}.

\bibitem{imagecompression}
\url{imagecompression.info}.

\bibitem{appendix}
\url{www.kecl.ntt.co.jp/people/wu.xiaomeng/icassp2020/appendix.pdf}.

\bibitem{AgaianSP07}
S.~S. Agaian, B.~Silver, and K.~A. Panetta,
\newblock ``Transform coefficient histogram-based image enhancement algorithms
  using contrast entropy,''
\newblock {\em {IEEE} Trans. Image Processing}, vol. 16, no. 3, pp. 741--758,
  2007.

\bibitem{ChenAPA06}
Z.~Chen, B.~R. Abidi, D.~L. Page, and M.~A. Abidi,
\newblock ``Gray-level grouping {(GLG):} {An} automatic method for optimized
  image contrast enhancement - {Part} {I:} {The} basic method,''
\newblock {\em {IEEE} Trans. Image Processing}, vol. 15, no. 8, pp. 2290--2302,
  2006.

\bibitem{WangMYWL15}
S.~Wang, K.~Ma, H.~Yeganeh, Z.~Wang, and W.~Lin,
\newblock ``A patch-structure representation method for quality assessment of
  contrast changed images,''
\newblock {\em {IEEE} Signal Process. Lett.}, vol. 22, no. 12, pp. 2387--2390,
  2015.

\bibitem{DabovFKE07}
K.~Dabov, A.~Foi, V.~Katkovnik, and K.~O. Egiazarian,
\newblock ``Image denoising by sparse {3-D} transform-domain collaborative
  filtering,''
\newblock {\em {IEEE} Trans. Image Processing}, vol. 16, no. 8, pp. 2080--2095,
  2007.

\bibitem{LiuTO13}
X.~Liu, M.~Tanaka, and M.~Okutomi,
\newblock ``Single-image noise level estimation for blind denoising,''
\newblock {\em {IEEE} Trans. Image Processing}, vol. 22, no. 12, pp.
  5226--5237, 2013.

\bibitem{LiuTO14}
X.~Liu, M.~Tanaka, and M.~Okutomi,
\newblock ``Practical signal-dependent noise parameter estimation from a single
  noisy image,''
\newblock {\em {IEEE} Trans. Image Processing}, vol. 23, no. 10, pp.
  4361--4371, 2014.

\end{thebibliography}
\end{document}